\documentclass[sigconf]{acmart}
\usepackage{balance}
\usepackage{eso-pic}
\usepackage{graphics}
\usepackage{float}
\usepackage{amsmath}
\usepackage{multirow}
\usepackage{caption}

\usepackage{subcaption}
\usepackage{booktabs}
\usepackage{graphicx}
\usepackage{wrapfig}
\usepackage{enumitem}
\usepackage{etoolbox}

\definecolor{lightgray}{gray}{0.9} 
\usepackage{colortbl}

\copyrightyear{2026}
\acmYear{2026}

\setcopyright{cc}
\setcctype{by-nc-nd}

\acmConference[KDD '26 PhD Consortium]{Proceedings of the 32nd ACM SIGKDD Conference on Knowledge Discovery and Data Mining}{August 9--13, 2026}{Jeju, Korea}

\acmBooktitle{Proceedings of the 32nd ACM SIGKDD Conference on Knowledge Discovery and Data Mining (KDD '26), August 9--13, 2026, Jeju, Korea}

\begin{document}

\settopmatter{authorsperrow=4}
\title{Towards a Data Flywheel for Embodied Intelligence in Logistics}

\author{Anlan Yu}
\affiliation{%
  \institution{Peking University}
  \city{Beijing}
  \country{China}
}
\email{yal6040@pku.edu.cn}

\author{Zaishu Chen}
\affiliation{%
  \institution{JD Logistics}
  \city{Beijing}
  \country{China}
}
\email{chenzaishu.1@jd.com}

\author{Zhiqing Hong}
\affiliation{%
  \institution{HKUST (Guangzhou)}
  \city{Guangzhou}
  \country{China}
}
\email{zhiqinghong@hkust-gz.edu.cn}

\author{Daqing Zhang}
\affiliation{%
  \institution{Peking University}
  \city{Beijing}
  \country{China}
}
\email{dqzhang@sei.pku.edu.cn}

\renewcommand{\shortauthors}{Anlan Yu et al.}

\begin{abstract}
Embodied intelligence is moving from laboratory demonstrations toward industrial deployment, with the logistics industry serving as a key application scenario. Learning-based policies offer a promising path beyond traditional perception-planning-control pipelines, but their scalability depends on how embodied data can be collected, organized, and reused. This research studies a data-centric framework for industrial embodied intelligence by constructing a logistics data flywheel. Our framework converts daily operations into reusable data assets, uses World Models to generate reliable supervision for long-tail parcel manipulation, and feeds deployment feedback back into policy improvement. As an initial result, \textit{WM-DAgger} introduces a World-Model-based data aggregation framework that synthesizes out-of-distribution recovery data for robust imitation learning. Building on this result, ongoing work explores how large-scale in-the-wild multimodal data, including labeled human demonstrations, unlabeled operational videos, and system-level robot logs, can be aligned for policy learning and transformed into feedback for continual system improvement.
\end{abstract}



\keywords{Industrial Data Mining; Data Flywheel; Robot Learning. }

\settopmatter{printfolios=true}
\maketitle


\section{Introduction}

Embodied intelligence is moving from laboratory toward industrial deployment. The logistics industry is a representative scenario~\cite{hong2025llm4har, hong2025experience}, where robots are expected to handle long-tail manipulation cases under reliability and throughput requirements. Recent advances in robot learning, including imitation learning~\cite{chi2025diffusion, fu2024mobile}, reinforcement learning~\cite{luo2025precise, tang2025deep}, and vision-language-action models~\cite{brohan2022rt, kimopenvla} suggest a promising path beyond traditional perception-planning-control pipelines. Rather than manually designing task-specific heuristic rules, end-to-end policies aim to acquire flexible and dexterous manipulation skills directly from data. The key to scaling this paradigm lies in the data scheme: how to collect, organize, generate, validate, and reuse embodied data at scale~\cite{lim2024open}.

Existing robot learning systems explore several data sources, including teleoperation~\cite{fu2024mobile}, UMI-style portable and embodiment-agnostic demonstrations~\cite{chi2024universal}, cross-embodiment robot datasets~\cite{lim2024open}, and unlabeled human videos~\cite{gao2026dreamdojo}. Meanwhile, World Models provide a complementary route for scaling embodied data by synthesizing additional training trajectories, and enabling learning from imagined interactions~\cite{hou2026world}. These efforts have improved the scalability of robot policy learning, but most existing data schemes are still built around laboratory-level demonstration tasks and curated benchmarks. While such data is usually well-structured and easy to align with policy learning objectives, its high collection cost and limited coverage of long-tail operational complexity remain key bottlenecks for scaling embodied agents in the real world.

Logistics industry provides a unique opportunity for data-centric embodied intelligence. In collaboration with JD Logistics, one of the largest logistics providers in China, this research is grounded in real industrial logistics scenarios. Unlike laboratory settings, real logistics sites continuously generate abundant and diverse embodied data, including human operation videos, labeled demonstrations, robot execution logs, failure cases, and system-level performance metrics. These data are naturally produced during daily operations, making them large-scale, diverse, and low-cost to collect. More importantly, they capture in-the-wild physical interactions under real operational constraints, spanning diverse parcel types, workspace layouts, and human manipulation strategies that are difficult to reproduce in controlled benchmark environments.

Realizing this opportunity raises three research questions.

\noindent \textbf{RQ1: How can legacy logistics automation data be converted into reusable embodied learning datasets?}
Industrial logistics sites already produce large amounts of data from traditional automation systems. Although these data were not originally collected for robot learning, they encode rich relationships among machine actions and world dynamics. The challenge is to align heterogeneous data into structured sequences that can support World Model training, VLA learning, and long-tail policy evaluation.

\noindent \textbf{RQ2: How can World Models generate reliable supervision for robust policy learning from limited demonstrations?}
Real demonstrations are costly and cannot cover the long tail of parcel manipulation, including grasp failures and recovery behaviors. World Models can synthesize additional trajectories beyond the limited state coverage of demonstrations, but generated data may contain hallucinated states, physically inconsistent transitions, or task-irrelevant actions. The challenge is to make World Models serve as reliable data engines that generate task-useful supervision.

\noindent \textbf{RQ3: How can logistics World Foundation Models be built and continuously improved through deployment feedback?}
Large-scale unlabeled logistics videos capture diverse parcel flows, human manipulation strategies, and complex physical interactions, while labeled demonstrations and robot rollouts provide action supervision at smaller scale. The challenge is to combine video-only pretraining, action-conditioned post-training, and deployment feedback into a continuous learning loop, so that World Models and robot policies can improve from both historical operational data and real deployment outcomes.

This research aims to address these RQs through an logistics industry embodied data flywheel. It connects operational data collection, World-Model-based data generation, and deployment feedback into a closed loop. As an initial result, \textit{WM-DAgger}~\cite{yu2026wm} demonstrates that World Models can synthesize reliable recovery trajectories for robust imitation learning. Ongoing work investigates how unstructured operational data can be aligned and transformed to support policy learning and continual improvement.
\section{World-Model-Driven Data Aggregation}

\begin{figure}[h]   
    \centering
    \includegraphics[width=0.47\textwidth]{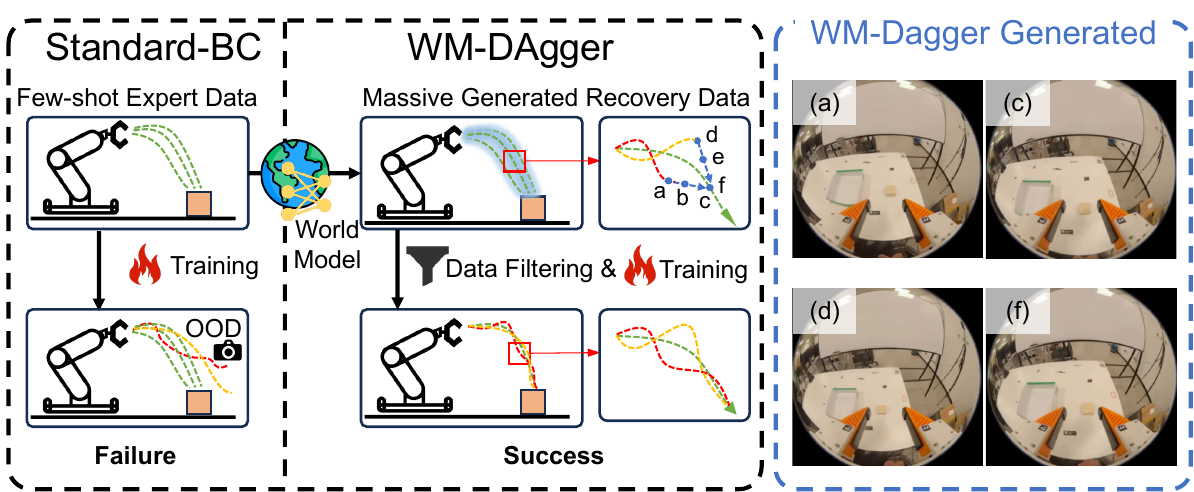}
    \caption{\textbf{WM-DAgger} mitigates the compounding errors of standard Behavioral Cloning (BC) by generating massive recovery supervision with a world model. (e.g., visual transitions $a \to b \to c$ and $d \to e \to f$). }
    \label{fig:teaser} 
\end{figure}

\noindent \textbf{Motivation and Challenges.}
Imitation learning often suffers from \textit{compounding execution errors}~\cite{ross2011reduction}: small prediction errors may drive the robot into out-of-distribution states, where the policy may further deviate towards failures. Existing data aggregation frameworks~\cite{ross2011reduction, kelly2019hg} mitigate this issue by asking human experts to provide recovery demonstrations during policy rollout, but such human-in-the-loop supervision is difficult to scale in industrial logistics, where data collection must be low-cost, repeatable, and compatible with large-scale deployment.

World Models provide a promising alternative by learning action-conditioned dynamics and predicting future observations~\cite{ding2025understanding}. This capability allows them to generate trajectories beyond the limited state coverage of real demonstrations, but the generated data may contain hallucinations, physically inconsistent transitions, or task-irrelevant actions. This motivates a concrete problem: how can World Models synthesize useful out-of-distribution data for imitation learning while avoiding misleading supervision? This problem is closely related to long-tail parcel manipulation, where robots may encounter failure cases that are difficult to fully cover with real demonstrations.

\noindent \textbf{Methodology.}
To address this problem, \textit{WM-DAgger}~\cite{yu2026wm} introduces a systematic World-Model-based data aggregation framework, as illustrated in Figure~\ref{fig:teaser}. It trains an action-conditioned World Model from limited demonstrations and exploratory interaction data. Given historical eye-in-hand observations and candidate robot actions, the World Model predicts future visual observations along potential recovery trajectories. The generated video-action trajectories are then aggregated with real demonstrations to train a more robust policy.

Synthetic recovery data is useful only when it is task-relevant and physically plausible for downstream policy learning. WM-DAgger therefore introduces two mechanisms. First, a \textit{Corrective Action Synthesis} module constructs task-oriented recovery trajectories by perturbing expert demonstrations toward nearby out-of-distribution states and then guiding the robot back toward the expert trajectory manifold. This prevents the generated actions from contradicting the task objective. Second, a \textit{Consistency-Guided Filtering} module removes generated trajectories whose terminal frames are visually inconsistent with corresponding real demonstration frames, reducing the impact of hallucinated or physically implausible rollouts before policy training.

\begin{table}[ht]
\caption{Summary of experimental data and results. (All tasks share 5 minutes of Play Data for world model initialization)}
\label{tab:main_results}
\centering
\resizebox{\columnwidth}{!}{%
\begin{tabular}{lcccccc}
\toprule
\multirow{2}{*}{\textbf{Evaluation Task}} & \multicolumn{2}{c}{\textbf{Dataset Size}} & & \multicolumn{3}{c}{\textbf{Success Rate (\%)}} \\
\cmidrule{2-3} \cmidrule{5-7}
& Real Demos & Gen. Data & & Std. BC & DMD & \textbf{Ours} \\
\midrule
Soft Bag Pushing (5-shot)  & 5  & 1500 && 26.7 & 40.0 & \textbf{93.3} \\
Soft Bag Pushing (20-shot) & 20 & 1500 && 30.0 & 56.7 & \textbf{96.7} \\
Pick-and-Place (Seen)      & 20 & 1500 && 11.1 & 32.2 & \textbf{84.4} \\
Pick-and-Place (Unseen)    & 20 & 1500 && 5.0  & 11.8 & \textbf{70.0} \\
Ballot Insertion           & 20 & 1500 && 13.3 & 26.7 & \textbf{73.3} \\
Towel Folding              & 20 & 1500 && 0.0  & 10.0 & \textbf{46.7} \\
\bottomrule
\end{tabular}%
}
\end{table}

\noindent \textbf{Results and Implications.}
Real-world experiments show that WM-DAgger can substantially improve few-shot imitation learning, as shown in Table~\ref{tab:main_results}. On a soft-bag pushing task related to logistics parcel handling, WM-DAgger achieves a 93.3\% success rate with only five expert demonstrations, significantly outperforming standard behavioral cloning and diffusion-based data augmentation baselines. The method also improves performance across multiple manipulation tasks involving rigid objects, deformable objects, and contact-rich interactions. 

This provides initial results for the broader scope of this research: World Models can serve as data engines for industrial embodied intelligence. More importantly, WM-DAgger shows that data generation and data validation must be considered jointly. In industrial settings, synthetic embodied data is valuable only when it leads to measurable improvements in downstream policy behavior under real deployment conditions. 
\section{Future Research Directions}

\textit{WM-DAgger} provides initial progress toward RQ2 by showing that World Models can generate reliable recovery supervision with task-aware validation. Future work will extend the logistics data flywheel in two directions.

\noindent \textbf{Legacy automation data as embodied datasets.}
To address RQ1, future work will transform data from traditional logistics automation systems into reusable learning datasets. Multi-sourced data will be aligned into observation-action-outcome sequences. Long videos will be segmented into interaction units, such as parcel arrival, clustering, and grasping, with lightweight metadata such as parcel type, workspace layout, and action outcome. The goal is to build an industrial logistics benchmark for World Model pretraining, action-conditioned prediction, weakly supervised representation learning, and long-tail policy evaluation.

\noindent \textbf{World Foundation Models with deployment feedback.}
To address RQ3, future work will pretrain logistics World Models from unlabeled operational videos and post-train them with labeled demonstrations and aligned robot or automation actions. The model will learn both general logistics dynamics and action-conditioned future prediction, supporting synthetic data generation, imagined rollouts, and World Action Models. Deployment feedback will close the loop by reusing successful rollouts for offline fine-tuning and failed rollouts for targeted recovery data generation and outcome-conditioned post-training.


\newpage
\bibliographystyle{ACM-Reference-Format}
\balance
\bibliography{survey}

\clearpage
\end{document}